\documentclass[conference]{IEEEtran}
\IEEEoverridecommandlockouts
\usepackage{cite}
\usepackage{amsmath,amssymb,amsfonts}
\usepackage{algorithmic}
\usepackage{graphicx}
\usepackage{textcomp}
\def\BibTeX{{\rm B\kern-.05em{\sc i\kern-.025em b}\kern-.08em
    T\kern-.1667em\lower.7ex\hbox{E}\kern-.125emX}}
\begin{document}

\title{Towards high-throughput 3D insect capture for species discovery and diagnostics
\thanks{This work was partly funded by CSIRO Transformational Biology Platform}
}

\author{
\IEEEauthorblockN{Chuong Nguyen, Matt Adcock, Stuart Anderson}
\IEEEauthorblockA{\textit{Quantitative Imaging}\\
\textit{CSIRO, Data61}\\
GPO Box 1700, Canberra ACT 2601, Australia \\
Chuong.Nguyen@csiro.au, Matt.Adcock@csiro.au, Stuart.Anderson@csiro.au}
\and
\IEEEauthorblockN{David Lovell}
\IEEEauthorblockA{\textit{Electrical Engineering \& Computer Science} \\
\textit{Queensland University of Technology}\\
Brisbane QLD 4001, Australia\\
David.Lovell@qut.edu.au}
\and
\IEEEauthorblockN{Nicole Fisher}
\IEEEauthorblockA{\textit{Australian National Insect Collection} \\
\textit{CSIRO, National Collections and Marine Infrastructure}\\
GPO Box 1700, Canberra ACT 2601, Australia \\
Nicole.Fisher@csiro.au}
\and
\IEEEauthorblockN{John La Salle}
\IEEEauthorblockA{\textit{Atlas of Living Australia}\\
\textit{CSIRO, National Collections and Marine Infrastructure}\\
GPO Box 1700, Canberra ACT 2601, Australia \\
John.Lasalle@csiro.au}
}

\maketitle

\begin{abstract}
Digitisation of natural history collections not only preserves precious information about biological diversity, it also enables us to share, analyse, annotate and compare specimens to gain new insights.
High-resolution, full-colour 3D capture of biological specimens yields color and geometry information complementary to other techniques (e.g., 2D capture, electron scanning and micro computed tomography). However 3D colour capture of small specimens is slow for reasons including specimen handling, the narrow depth of field of high magnification optics, and the large number of images required to resolve complex shapes of specimens. In this paper, we outline techniques to accelerate 3D image capture, including using a desktop robotic arm to automate the insect handling process; using a calibrated pan-tilt rig to avoid attaching calibration targets to specimens; using light field cameras to capture images at an extended depth of field in one shot; and using  3D Web and mixed reality tools to facilitate the annotation, distribution and visualisation of 3D digital models. 

\end{abstract}


\section{Introduction}
Specimen collections are of most scientific value when they are readily available for further study, analysis and annotation---hence increasing interest and adoption of digitisation.
In entomology, common digitisation methods include high-resolution mosaic images of insects trays \cite{mantle2012whole} or stacked multifocus images \cite{brecko2014focus}. While these techniques capture color information, they do not capture 3D geometry of specimens. Micro Computed Tomography (Micro CT) captures 3D geometry \cite{faulwetter2013micro}, but not true color information, and scanning is laborious and expensive \cite{Cross2017}. An alternative solution is to use gray-scale video from Rotational Scanning Electron Micrographs (rSEM) \cite{cheung2013rotational} to show 3D geometry at a fixed tilt angle without performing 3D reconstruction. These mainstream solutions have been excellent tools for researchers but are difficult to scale up to scan collections of tens of million specimens.

Recent advancements in 3D photogrammetry for small specimens promise a new scalable way to create high quality 3D mesh models at low cost. Nguyen et al. \cite{nguyen2014capturing} proposed a 3D full-color scanning system that captures multi-view multi-focus image-sets of insects of sizes from 3mm and larger to generate 3D mesh models. The system demonstrated that it is possible to obtain high resolution full color 3D mesh insect models using off-the-shelf components. Brecko et al. \cite{brecko2014focus} proposed an effective low-cost setup including an Ikea closet and produced a detailed 3D mesh model of a large \textit{Dicranorrhina} sp. beetle using Agisoft Photoscan software. ZooSphere \cite{ZooSphere2017} is an initial effort to capture similar multiple view images of large number of insect specimens and share image data for future 3D reconstruction. 
To an extreme end, Martins et al. \cite{martins2015r2obbie} proposed a sophisticated system with an industrial robotic arm carrying a camera and a LED array to capture both multi-view and photometric stereo data. 

While these techniques enable realistic 3D model capture, throughput is low: image capture is time-consuming and the significant human input is required. 
To accelerate the throughput of 3D photogrammetry for small specimens, this paper proposes a number of strategies to deal with bottlenecks across the digitization process. 

\section{Identified processes and solutions}

\subsection{Automatic specimen handling}
\begin{figure}[htbp]
\centerline{\includegraphics[width=0.7\columnwidth]{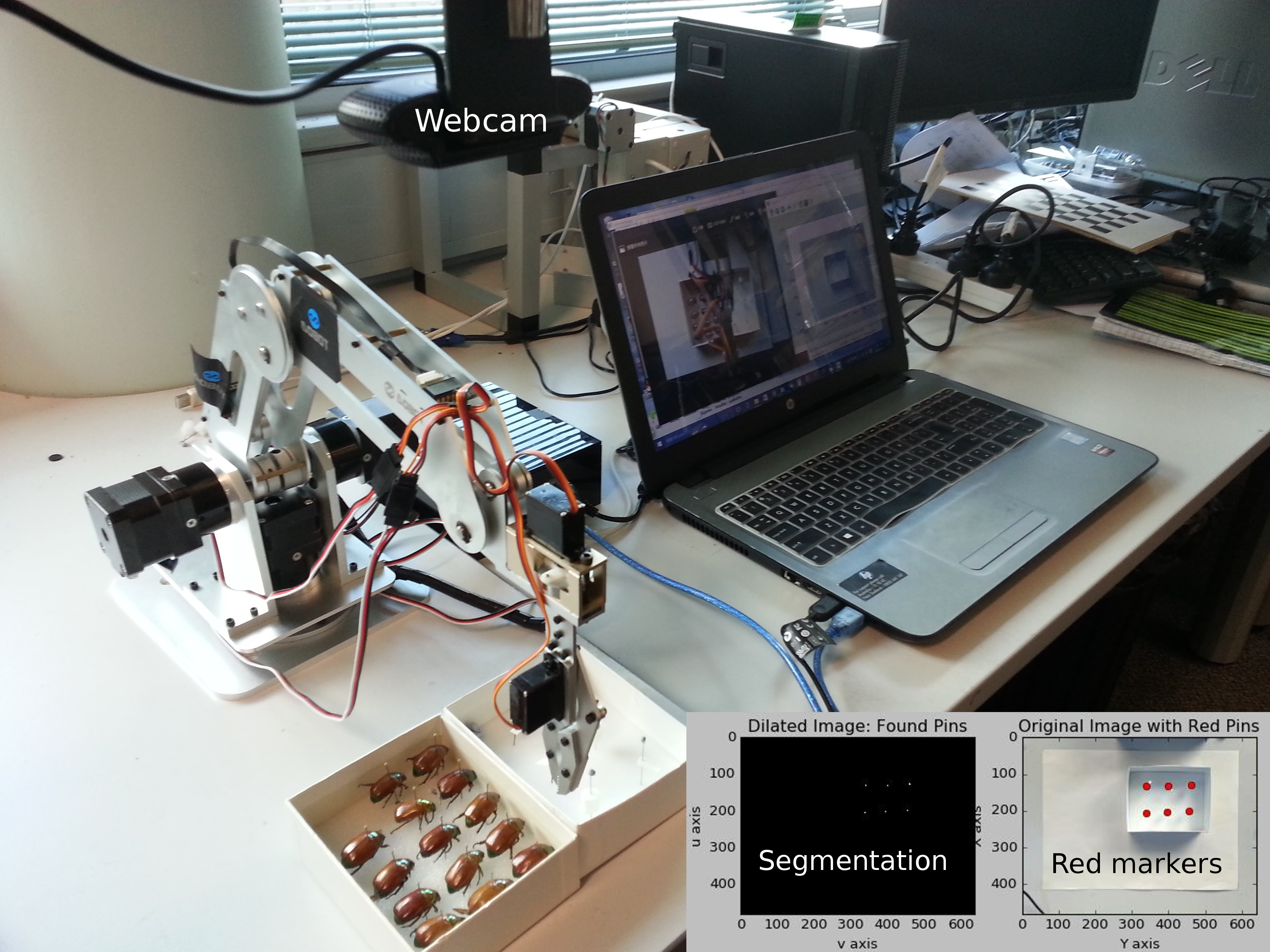}}
\caption{Simple desktop robot setup with top-view camera for insect manipulation. Inset shows red markers to enhance x-y position detection of pin-heads.}
\label{fig:Dobot}
\end{figure}

Emerging low-cost, but precise desktop robot arms (e.g., Dobot \cite{Dobot2017}) can be adapted to automate the loading of specimens from tray to 3D scanner and back. Fig. \ref{fig:Dobot} shows simple setup from a prototype system \cite{BoyaYu2016} where a calibrated top-view camera provides pin heads' x-y coordinates for the robotic arm. More sophisticated robotic arms can be developed for both picking specimens from a tray, and rotating them in front of a camera or camera array for multi-view image capture.

\subsection{Target-less multi-view image capture for 3D reconstruction}
Coded patterns or targets are usually used during image capture to help estimate camera pose and scale factor. Without them, camera pose estimation is unreliable and largely dependent on the shape and texture of the specimen. However the smaller the specimen, the harder it is to print and attach a pattern of similar size---Nguyen et al. \cite{nguyen2014capturing} required a high-resolution printer and a skilled practitioner to attach a target to a specimen under a microscope. To eliminate the need for such patterns, a precise pan-tilt rig can be used and calibrated so that camera poses can be computed directly from the readings of the rig's motors. Such a system could be built from off-the-shelf components such as those in the Cognisys StackShot 3X \cite{Cognisys2017} which has an angular resolution of 0.01 degree. We have successfully implemented a software interface to control StackShot 3X with one or more tethered cameras.

\subsection{Speeding up image acquisition}
Image acquisition can be time consuming. While the DSLR camera in \cite{nguyen2014capturing} was automated, its speed was limited by its mechanical shutter. High-resolution and large sensor size cameras (e.g.,  Grasshopper3 \cite{PointGrey2017}) can be used to accelerate image acquisition. In addition, synchronized multiple cameras can further increase image acquisition speed. Camera arrays can also reduce or even eliminate rotation motion.

An additional constraint for tiny specimens is the small depth of focus of high magnification lenses: multiple, partially-focused images are taken at incremental depth positions and processed (or stacked) to produce an all-in-focus image of a specimen. Not only is this multi-focus image capture and  processing time consuming,  the resulting image can contain significant artifacts and distortion leading to unsuccessful 3D reconstruction. We propose using a light field camera (e.g., Lytro Illum) to acquire a single-shot extended depth of field image (although with a reduced resolution). An additional macro lens could be used to achieve the desired magnification.

\subsection{3D web-based data augmentation and publication}
Processing and deploying 3D data from a large collection of specimens can be a significant bottleneck. 3D mesh models need to be reconstructed, possibly edited to remove noise and errors, and annotated to add semantic information. 
Existing 3D editing tools designed for CAD or 3D Animation involve many manual steps and are therefore unsuitable for non-experts. 
Platforms for uploading 3D models to the web are emerging, such as Sketchfab \cite{Sketchfab2017}, but remain focussed on display only. Augmented Reality (AR) technologies, such as Microsoft HoloLens, Google Tango, and even smartphones, are promising as devices for interaction and annotation but remain difficult to use or share across devices of different makers. Our current work \cite{anderson2015towards} includes the extension of 3D Web applications into AR to allow multiple AR to connect to and work on specimens in the same virtual environment.

\section{Discussion and conclusion}
Recent developments in 3D digitization of insect collections are reviewed, and demonstrate that 3D photogrammetry for small specimen is opening up a major opportunity for low-cost but meaningful solutions for large scale digitization. We identify four different areas that cause bottlenecks for a high throughput digitization system. Finally we propose practical solutions to eliminate or reduce the bottlenecks through presenting some of our early experience in pursuing these solutions. 

\bibliographystyle{IEEEtran}
\bibliography{IEEEabrv,Hight_Throughput_3D_Insect_Digitizer}

\end{document}